\documentclass[11pt,a4paper]{article}

\usepackage[utf8]{inputenc}
\usepackage{amsmath,amssymb}

\usepackage{graphicx}
\usepackage{float}          
\usepackage{multirow}       

\usepackage[margin=1in]{geometry}

\usepackage{etoolbox}

\patchcmd{\abstract}{\quotation}{\small\quotation}{}{}
\patchcmd{\abstract}{\list}{\small\list}{}{}

\usepackage[font=small,labelfont=bf]{caption}
\usepackage[colorlinks=true,linkcolor=black,citecolor=black,urlcolor=black]{hyperref}

\usepackage{cite}

\title{Softly Induced Functional Simplicity:\\ Implications for Neural Network Generalisation, Robustness, and Distillation}
\author{Maciej Glowacki \\ European Organization for Nuclear Research (CERN) CH-1211 Geneva 23, Switzerland \\ \texttt{maciej.glowacki@cern.ch}}

\date{}  

\begin{document}
\maketitle
\begin{abstract} 

Learning robust and generalisable abstractions from high-dimensional input data is a central challenge in machine learning and its applications to high-energy physics (HEP). Solutions of lower functional complexity are known to produce abstractions that generalise more effectively and are more robust to input perturbations. In complex hypothesis spaces, inductive biases make such solutions learnable by shaping the loss geometry during optimisation. In a HEP classification task, we show that a soft symmetry respecting inductive bias creates approximate degeneracies in the loss, which we identify as pseudo-Goldstone modes. We quantify functional complexity using metrics derived from first principles Hessian analysis and via compressibility. Our results demonstrate that solutions of lower complexity give rise to abstractions that are more generalisable, robust, and efficiently distillable.

\end{abstract}

\noindent
\textbf{\textsf{Keywords:}} Particle physics, Machine Learning, Generalisation, Interpretability, Loss geometry, Robustness, Model distillation

\section{Introduction}

Machine learning models are increasingly used in high-energy physics (HEP), particularly in collider experiments, for tasks such as object reconstruction, event classification, and anomaly detection~\cite{MLinHEP}. Success in these tasks depends on a model’s ability to extract abstractions from high-dimensional, noisy inputs that generalise reliably to unseen data. Such abstractions emerge when inputs are mapped to lower dimensional representations that capture salient structure. Fundamentally, for data to be meaningfully compressed, it must first be understood. While compression is central to representation learning~\cite{shwartz-ziv2023compress}, we argue that functional complexity plays a complementary role. Whereas compression reduces the volume of information, simplicity biases ensure that this information is encoded through smooth, low-curvature mappings by guiding the optimiser toward low complexity solutions in the loss geometry. These regions are empirically associated with improved generalisation and robustness under variations in the input distribution~\cite{andrew}. In this work, we compare learned solutions obtained with and without a soft symmetry-constraining inductive bias, and quantify their functional complexity using geometry diagnostics and compressibility. We find that soft biases reduce the solution's effective dimensionality by inducing low-curvature trajectories. We further demonstrate that reductions in complexity are associated with improved robustness, out-of-distribution generalisation, and distillability in a realistic HEP classification task.

\section{Conceptual Framework}

\subsection{Theory of Generalisation}

Understanding why and how neural networks generalise remains a central question in learning theory. In this work, we adopt the viewpoint that generalisation is closely linked to the functional complexity of the learned solution. This is partly because solutions of lower functional complexity tend to occupy a larger volume in parameter space, which under PAC‑Bayes intuitions can lead to a tighter bound on the expected generalisation error~\cite{pac}. We quantify functional complexity in two ways.

Firstly, through the effective dimensionality, estimated from the Hessian of the loss. For a Hessian $\mathcal{H}$ evaluated at network parameters $\mathbf{w}$,
\begin{equation}
\mathcal{H}_{ij} = \frac{\partial^2 \mathcal{L}(\mathbf{w})}{\partial w_i \, \partial w_j},
\end{equation}
the effective dimensionality can be approximated by its trace, $\mathrm{Tr}(\mathcal{H})$. Large trace values indicate many sharp directions in the loss geometry, whereas smaller values correspond to flatter minima with fewer sensitive directions in parameter space. While flatness alone is not a complete explanation for generalisation, reductions in effective dimensionality are empirically associated with improved performance~\cite{flatness}.

Secondly, we quantify functional complexity through compressibility, operationalised by how efficiently a trained model can be distilled into a fixed, lower-capacity student. For a given student architecture and optimisation procedure, differences in distillation performance reflect differences in the complexity of the teacher function. Since models with identical parameter counts can exhibit different distillation behaviour, parameter count alone is not a sufficient proxy for functional complexity.

\begin{figure}[h!]
    \centering
    \includegraphics[width=0.8\textwidth]{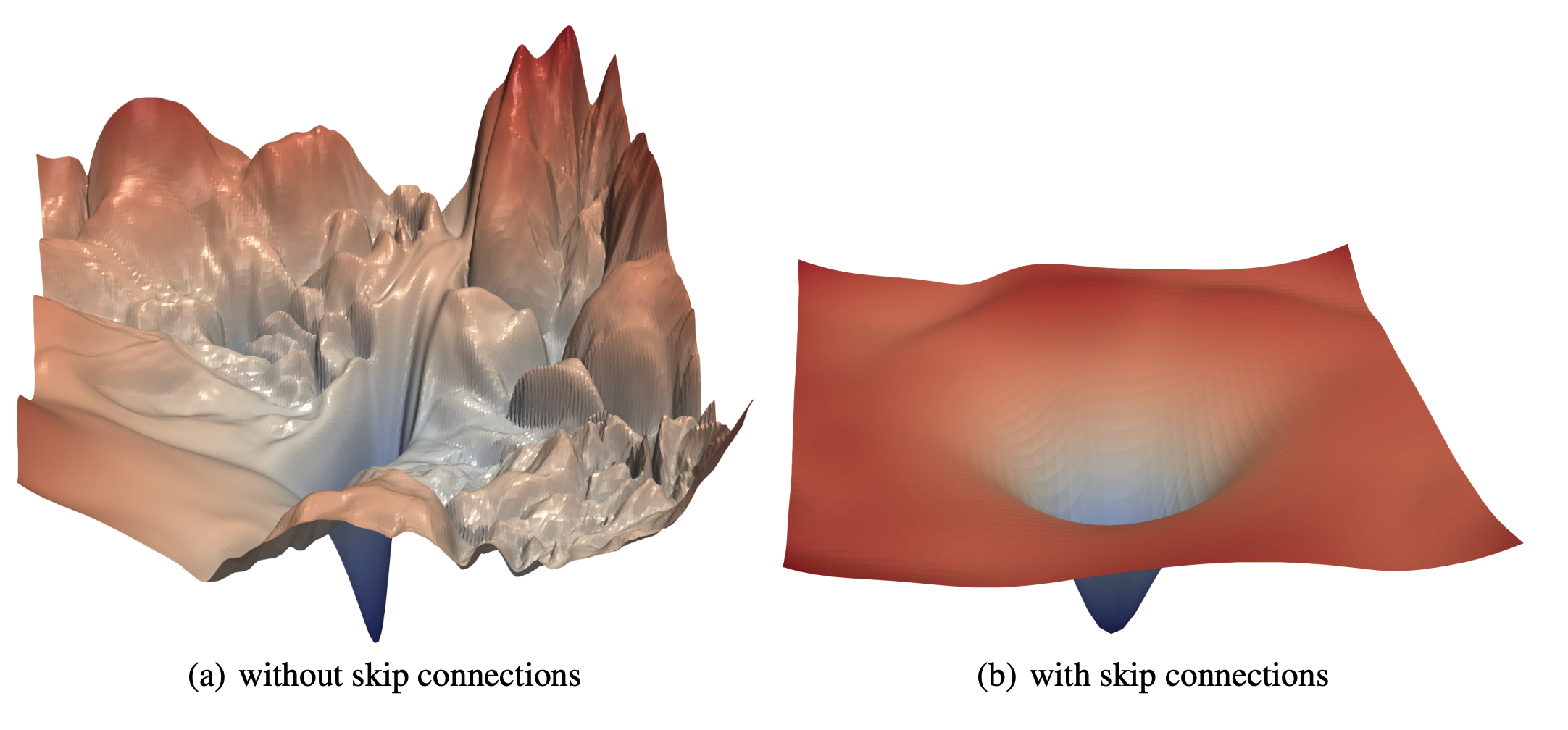}
    \caption{Visualisation of loss geometry of ResNet-56 with skip connection (left) versus without (right). Wider minima correspond to reduced effective dimensionality. Figure taken from~\cite{visualisaing_loss}.}
    \label{fig:minimas}
\end{figure}

Moreover, recent work has highlighted connections between loss geometry and the robustness of learned solutions~\cite{hessianconnection}. The leading Hessian eigenvector corresponds to the parameter direction of largest local curvature and is often aligned with perturbations that strongly deform the decision boundary in input space. Minima with a smaller leading eigenvalue are flatter along this dominant direction and are empirically associated with smoother decision boundaries and larger effective margins~\cite{hessianconnection}. Together with measures of overall effective dimensionality, the leading eigenvalue of the loss Hessian provides a diagnostic that relates loss geometry with robustness to input perturbations.

\begin{figure}[h!]
    \centering
    \includegraphics[width=0.9\textwidth]{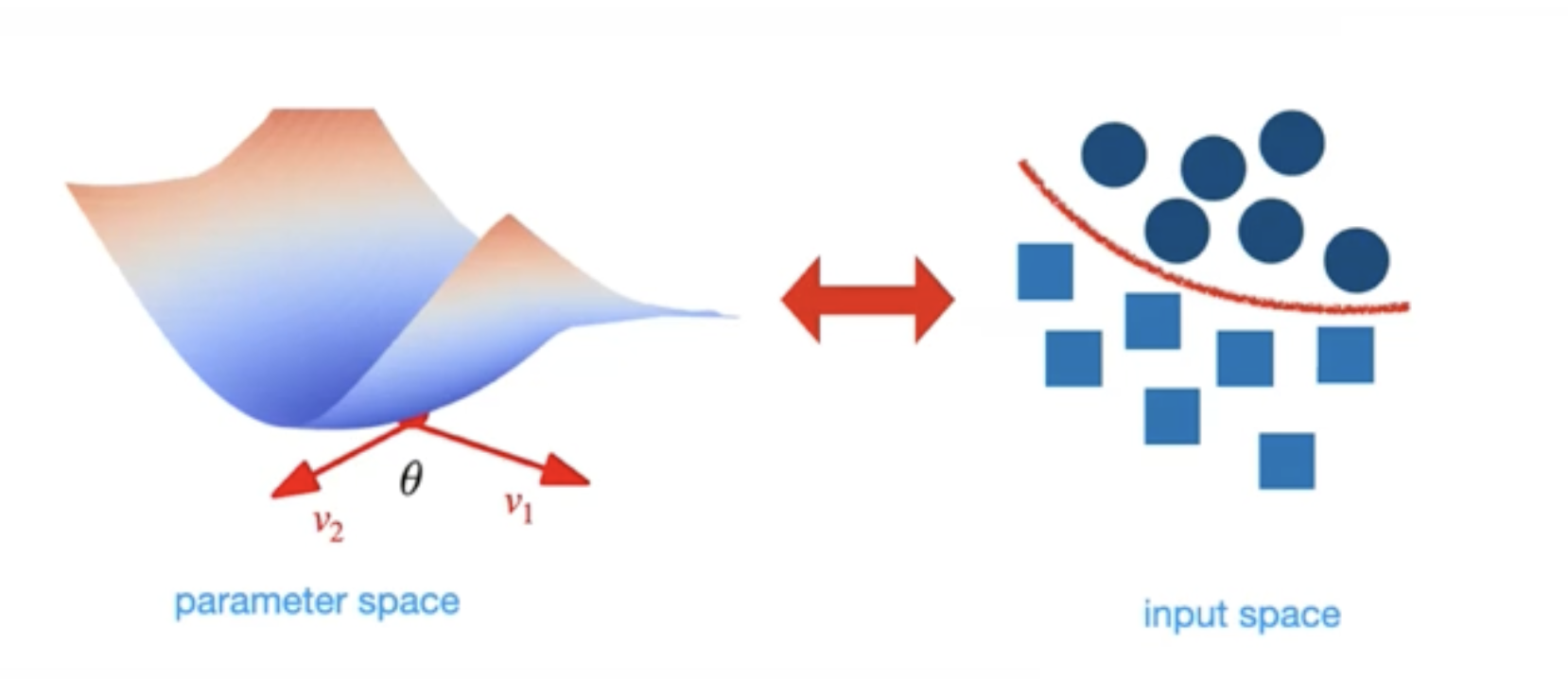}
    \caption{Illustrating the connection between parameter-space curvature (left) and effective margins in input space (right). Leading Hessian eigenvectors correspond to directions that deform the decision boundary, while the number of significant modes controls the number of independent deformation directions.}
    \label{fig:param-function-space}
\end{figure}

\subsection{Simplicity Bias}

Many deep learning techniques and architectures introduce an implicit simplicity bias into optimisation, reshaping the loss geometry to make it traversable. These biases can be hard or soft; hard biases impose architectural constraints that restrict the accessible hypothesis space, which can be overly restrictive~\cite{harmful}. While soft biases preserve expressivity but steer optimisation toward simpler solutions within a flexible model class. Both forms generally reduce the effective dimensionality of the solution and suppress leading Hessian eigenvalues.

\begin{figure}[h!]
\centering
\includegraphics[width=1.0\textwidth]{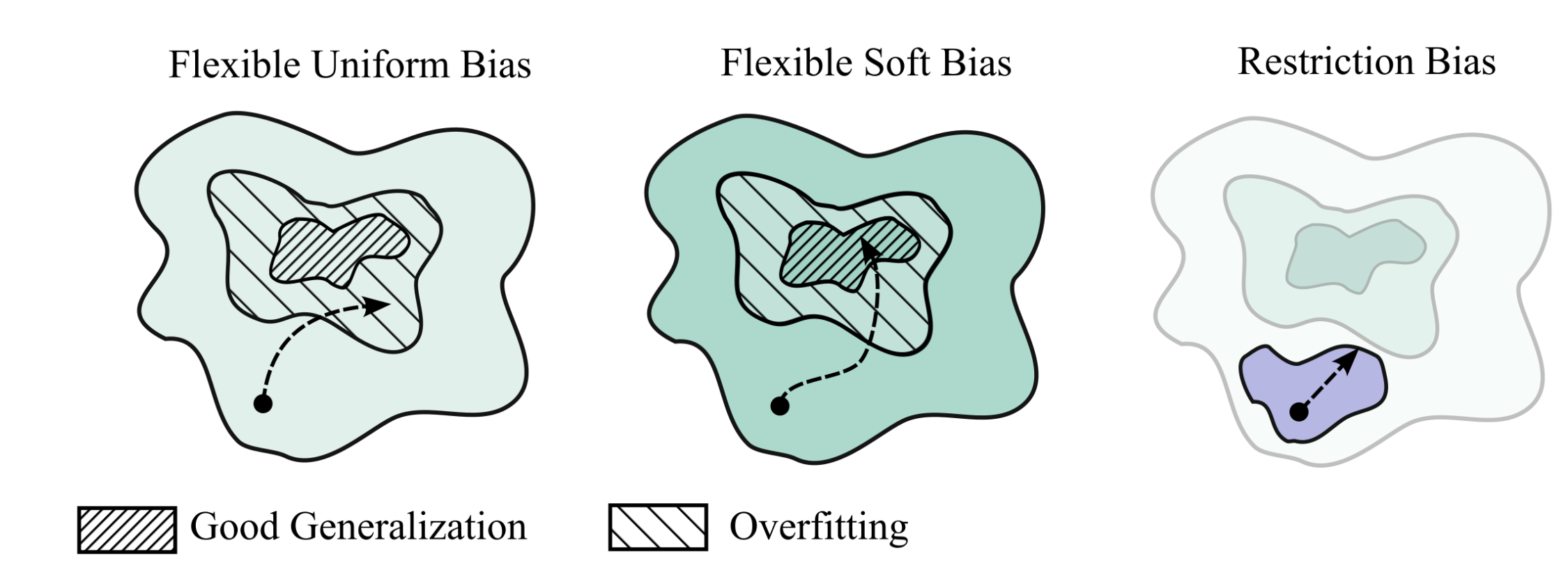}
\caption{Hard and soft inductive biases and their effect on optimisation and generalisation. 
Without inductive bias (left), optimisation explores a large hypothesis space with no preference among solutions that fit the data, often leading to overfitting. 
Soft inductive biases (middle) preserve expressivity while biasing optimisation toward preferred regions of the loss geometry associated with better generalisation. 
Hard inductive biases (right) restrict the hypothesis space to enforce desired properties, reducing overfitting at the cost of reduced expressivity. 
Figure taken from~\cite{andrew}.}
\label{fig:soft-hard-bias}
\end{figure}

Soft inductive biases are typically implemented through regularisation terms, auxiliary constraints in the training objective, or skip connections. Most modern training techniques incorporate several soft biases simultaneously. Canonical examples include:
\newpage

\begin{itemize}

    \item \textbf{Weight Decay ($L_2$ Regularisation)} penalises large weights, effectively acting as a Gaussian prior over network parameters and biasing optimisation toward low-norm solutions~\cite{bishop2006prml}. For parameter $\boldsymbol{\theta}$, the gradient update becomes
\begin{equation}
    \boldsymbol{\theta}_{t+1} = \boldsymbol{\theta}_t - \eta \left( \nabla_{\boldsymbol{\theta}} \mathcal{L} + \lambda \boldsymbol{\theta}_t \right),
\end{equation}
which is equivalent to optimising the posterior probability under
$p(\boldsymbol{\theta}) \propto \exp(-\lambda \|\boldsymbol{\theta}\|^2)$. By penalising large weights, weight decay suppresses sharp directions in the loss geometry. This reshaping steers optimisation toward solutions of lower functional complexity.

    \item \textbf{Skip Connections} update the hidden state $h_k$ at layer $k$ via
    \begin{equation}
        h_{k+1} = h_k + f_k(h_k),
    \end{equation}
    where $f_k$ is the layer transformation. This results in the corresponding layer-wise Jacobian,
    \begin{equation}
        J_k = \frac{\partial h_{k+1}}{\partial h_k} = I + \frac{\partial f_k}{\partial h_k},
    \end{equation}
    that has eigenvalues clustered near unity, mitigating vanishing and exploding gradients and biasing optimisation toward smooth, low-curvature trajectories that yield solutions of lower functional complexity~\cite{he2016deep}. The effect of skip connections on the loss geometry is visualised in Figure~\ref{fig:minimas}.

\end{itemize}

\subsection{Symmetries}

Symmetries provide a structured and physically motivated form of simplicity bias by constraining how functions respond to transformations of their inputs. For a symmetry group $G$ acting on an input space $X$, a function $f: X \to Y$ is invariant if $f(g \cdot x) = f(x)$ for all $g \in G$, and equivariant if $f(g \cdot x) = \rho(g) f(x)$, where $\rho(g)$ specifies a known transformation of the output. Convolutional Neural Networks (CNNs), for example, implement a hard inductive bias via kernel weight sharing to enforce translation equivariance~\cite{CNN}. In physics, symmetry constraints applied to theories have yielded tremendous progress and form the basis of the Standard Model~\cite{peskin1995introduction}. Lorentz symmetry expresses the invariance of fundamental interactions under rotations and boosts in space-time~\cite{Lorentz}. While collider interactions are Lorentz invariant at the parton level, this symmetry is broken in observed data~\cite{laGTR} by detector effects, finite acceptance, and reconstruction techniques. Enforcing strict symmetry constraints can therefore be overly restrictive for realistic collider tasks. Several approaches implement Lorentz symmetry as a hard inductive bias through invariant or equivariant architectures~\cite{Lorentz, laGTR, loca}. Alternatively, symmetry can be enforced softly by augmenting the loss with a term that penalises symmetry violations, as in SEAL~\cite{SEAL}, steering the model toward approximately symmetric solutions while preserving expressivity.

\section{Method}

This section describes the experimental setup used to evaluate the impact of functional complexity on models trained for a HEP classification task. Specifically, models are trained to classify final state particles originating from the decay and hadronization of initial quarks, bosons, or gluons, which are clustered into jets~\cite{jets}.  

To encourage functional simplicity during optimization, we adopt the SEAL methodology~\cite{SEAL}. Specifically, we define a second view of the jet by applying random 3D Lorentz boosts to all particles. We introduce a soft inductive bias by adding a regularisation term to the objective function that penalizes the Mean Squared Error (MSE) between the model outputs for these two views. All group actions and regularization details follow exactly the procedure in~\cite{SEAL}. Accordingly, we consider two model variants:

\begin{itemize}
    \item \textbf{Unconstrained model}: optimises only the binary cross-entropy (BCE) loss.
    \item \textbf{Constrained model}: optimises
    \[
        \mathcal{L} = \mathcal{L}_{\mathrm{BCE}} + \lambda \, \mathcal{L}_{\mathrm{SEAL}},
    \]
    with $\lambda = 0.1$.
\end{itemize}

All other training hyperparameters are held fixed to isolate the effect of the inductive bias.

\subsection{Dataset}
We use the publicly available \texttt{hls4ml} jet dataset~\cite{hls4ml_dataset}, which contains particle jets initiated by gluons, quarks, and bosons. Each jet consists of up to 32 constituent particles, sorted by their transverse momentum ($p_T$), with raw four-momentum components $(E, p_x, p_y, p_z)$. For each constituent, we construct a feature vector from four-momentum derived quantities and normalised relative to the parent jet:
\[
\Delta \eta,\, \Delta \phi,\, \log(p_T / p_T^\text{jet}),\, 
\log p_T,\, \log(E / E^\text{jet}),\, \log E,\, \Delta R
\]
where $\Delta \eta$ and $\Delta \phi$ are the differences in pseudorapidity and azimuthal angle with respect to the jet axis, and $\Delta R = \sqrt{(\Delta \eta)^2 + (\Delta \phi)^2}$ measures the angular distance from the jet axis. This feature set captures both local constituent-level structure and global jet properties. The dataset is divided into 200,000 jets for training, 49,560 for validation, and 104,704 for testing. All reported experiments use the test set for evaluation.

\subsection{Model Architecture}

All studies employ a transformer encoder operating on sets of jet constituents. Each constituent is first projected from the input dimension to an embedding of size $d_{\mathrm{model}} = 256$ using a linear layer followed by a ReLU activation. The embedding is processed by a stack of three transformer encoder layers. Each encoder layer consists of 4 attention heads, a feed-forward network with hidden dimension 256, skip connections, layer normalization, and ReLU activations. After the encoder stack, constituent embeddings are compressed by first applying mean pooling along the constituent dimension, followed by dimensionality reduction from 256 to 128 through three linear layers with ReLU activations. The network outputs a single scalar logit for binary classification. The model contains a total of 1,256,356 trainable parameters.

\subsection{Training and Evaluation Procedure}
\label{sec:training}
Both models are trained for $100$ epochs using the Adam optimiser with a fixed learning rate of 0.005 and weight decay of 0.0005 to stabilise training from which the model with the lowest validation loss is chosen. We train three independent models per configuration with different random initialisations. For standard evaluation, reported metrics and uncertainties correspond to averages over these three seeds. For computationally expensive analyses, including Hessian computation, distillation, and robustness studies, we select only the model achieving the lowest validation loss from each triplet. All experiments are implemented in \texttt{PyTorch}.

\newpage
\subsection{Hessian Analysis}
For each model, we compute the Hessian of the loss with respect to the network parameters and extract its dominant eigenpair $(\nu_1, \lambda_1)$ as well as the trace $\mathrm{Tr}(\mathcal{H})$. All quantities are computed using the \texttt{pyhessian} package, which employs stochastic Lanczos iterations to estimate the top Hessian eigenpairs. The trace is estimated using the Hutchinson estimator~\cite{pyhessian}.

\subsection{Distillation Setup}
To assess the functional complexity through compressibility of the learned solutions, we distill each trained teacher model (unconstrained and constrained) into an identical, lower-capacity student network with fewer parameters. The student has a total of 487,665 trainable parameters and is trained to match the teacher’s behaviour by minimising the mean squared error (MSE) between their output logits,
\[
    \mathcal{L}_{\mathrm{distill}}
    = \frac{1}{B}\sum_{i=1}^B 
    \| s(x_i) - t(x_i) \|_2^2 ,
\]
where $t(x)$ and $s(x)$ denote the teacher and student outputs, respectively. The student architecture and optimisation procedure are held fixed across both distillation experiments, ensuring that differences in distillation performance directly reflect differences in the functional complexity of the teacher models.

\section{Experiments}

\subsection{In-Distribution Performance}

\begin{figure}[H]
    \centering
    \includegraphics[width=0.5\textwidth]{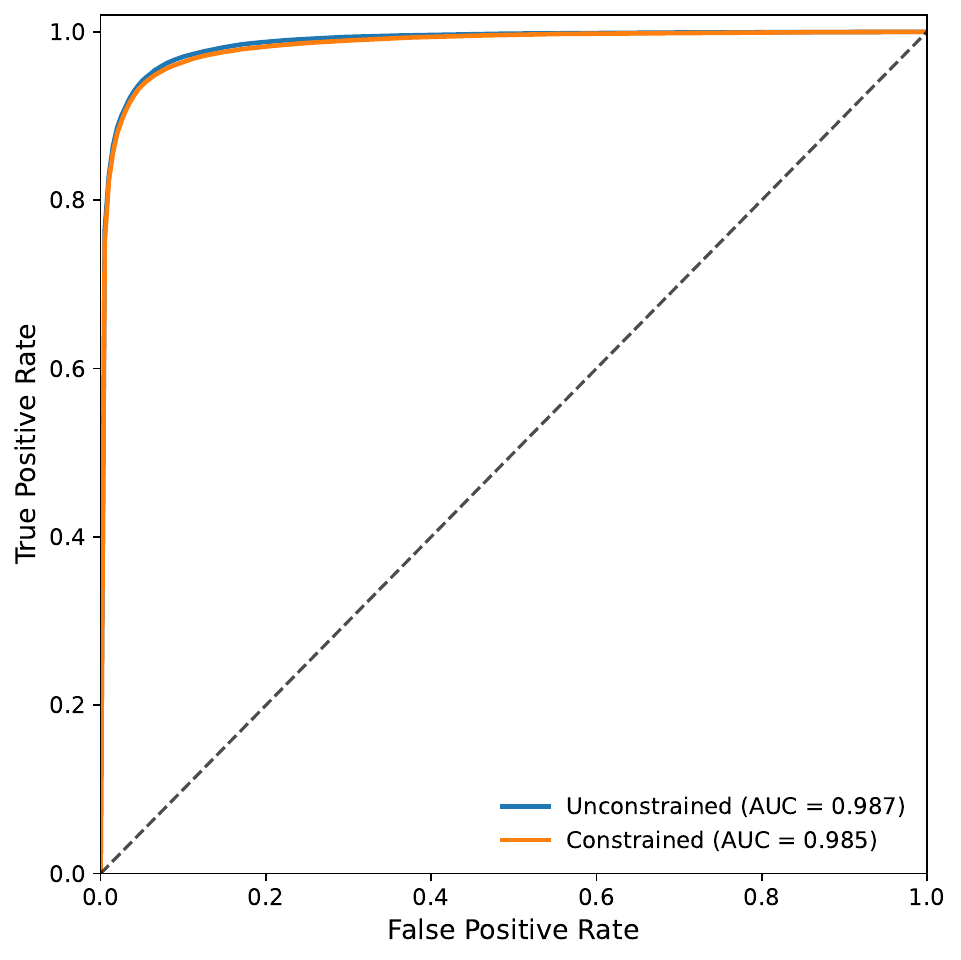}
    \caption{Receiver operating characteristic (ROC) curves for the in-distribution Gluon vs W boson classification task, comparing unconstrained and constrained models. Both models achieve comparable discrimination performance.}

    \label{fig:performance-summary}
\end{figure}

We first evaluate the effect of the soft inductive bias on in-distribution performance. Both constrained and unconstrained models are trained to discriminate gluon-initiated jets from decaying $W$ bosons, following the procedure in Section~\ref{sec:training}. Table~\ref{tab:indist-acc} reports the accuracy and the area under the receiver operating characteristic curve (ROC-AUC), with the corresponding ROC curves shown in Figure~\ref{fig:performance-summary}. Differences between models are negligible, indicating that the soft inductive bias does not impair nominal classification performance. The unconstrained variant exhibits slightly larger uncertainties, which could indicate marginally higher variability in convergence across seeds, though the effect is minimal.

\begin{table}[H]
    \centering
\caption{In-distribution classification performance of unconstrained and constrained models on the Gluon vs W boson task. Values are averaged over three independent runs; uncertainties indicate standard deviation across runs.}

    \vspace{0.3cm}
    \begin{tabular}{l l c c}
        \hline
        Task & Model & Accuracy & ROC-AUC \\
        \hline
        \multirow{2}{*}{Gluon vs W boson} 
            & Unconstrained & $0.9424 \pm 0.0024$ & $0.9875 \pm 0.0005$ \\
            & Constrained   & $0.9400 \pm 0.0009$ &  $0.9850 \pm 0.0002$\\
        \hline
    \end{tabular}
    \label{tab:indist-acc}
\end{table}

Across Hessian-based metrics, the constrained model exhibits both a smaller leading eigenvalue $\lambda_1$ and a reduced Hessian trace $\mathrm{Tr}(\mathcal{H})$ compared to the unconstrained baseline. Table~\ref{tab:hessian-ratios} reports the ratios of the leading eigenvalue and Hessian trace between the constrained and unconstrained models. In both cases, ratios below unity indicate that the soft inductive bias consistently reshapes loss geometry to drive optimisation toward solutions of lower effective dimensionality and deformation.

\begin{table}[H]
    \centering
    \caption{Ratios of Hessian-based curvature metrics between the symmetry-constrained and unconstrained models. Values below unity indicate reduced curvature and effective dimensionality for the constrained model.}
    \label{tab:hessian-ratios}
    \begin{tabular}{l c}
        \hline
        Metric & Value \\
        \hline
        $\lambda_1^{\mathrm{con}} / \lambda_1^{\mathrm{unc}}$ &  0.079 \\
        $\mathrm{Tr}(\mathcal{H})^{\mathrm{con}} / \mathrm{Tr}(\mathcal{H})^{\mathrm{unc}}$ & 0.119  \\
        \hline
    \end{tabular}
\end{table}

\begin{figure}[h!]
\centering
\includegraphics[width=0.49\textwidth]{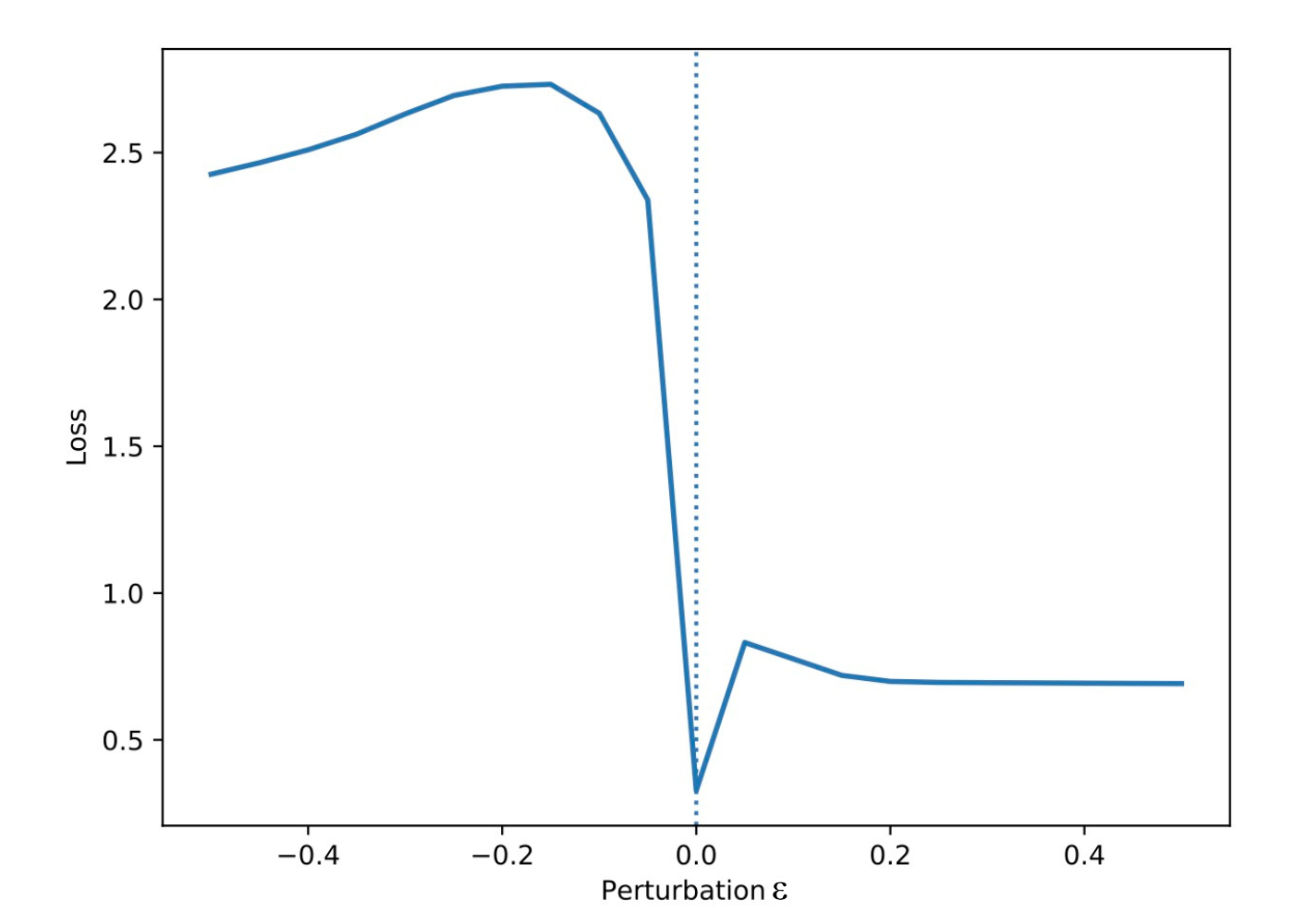}
\includegraphics[width=0.49\textwidth]{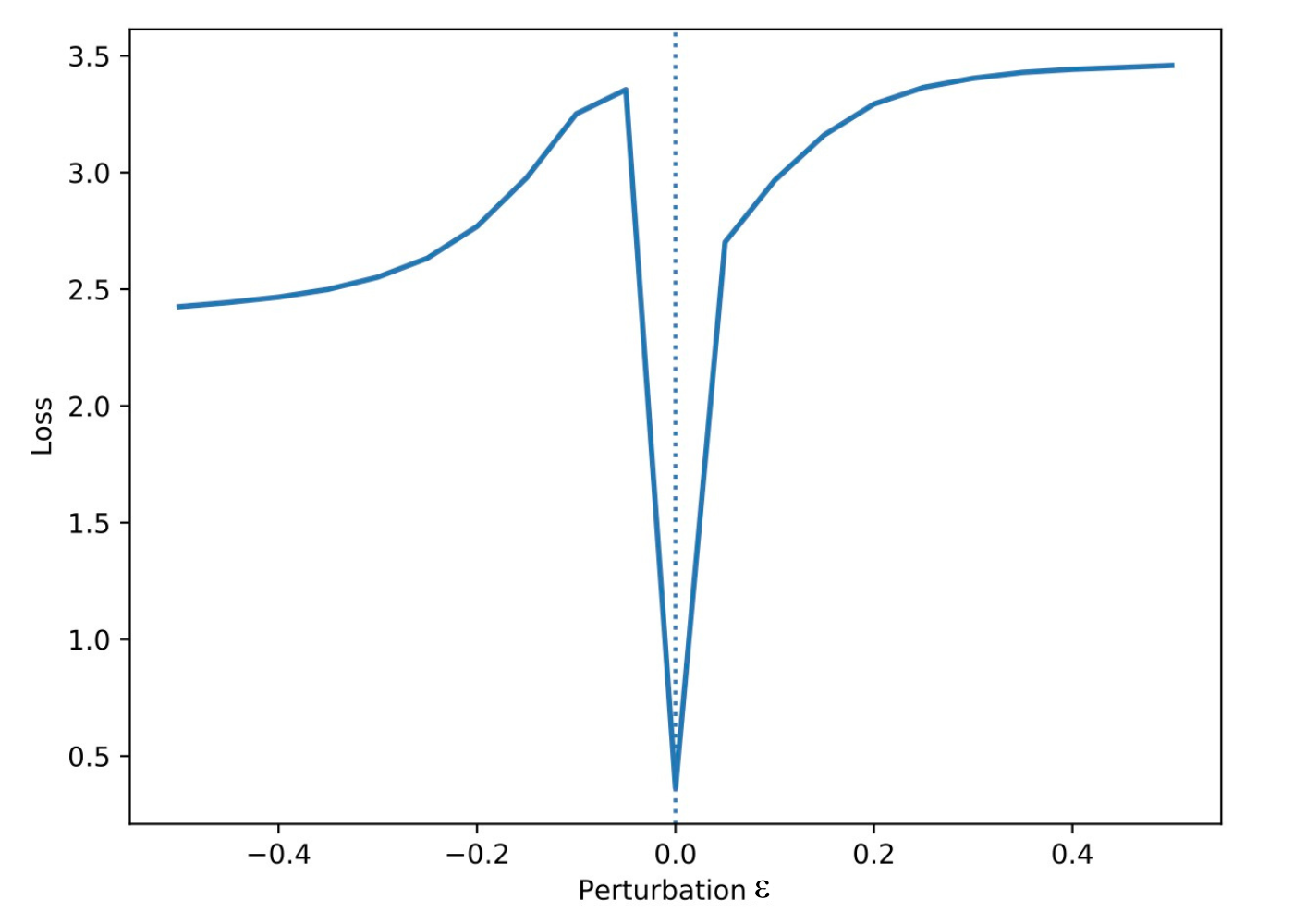}
\caption{Loss geometry along the leading Hessian eigenvector $\nu_1$ for the constrained (left) and unconstrained (right) models. 
The constrained model exhibits a pseudo-Goldstone mode aligned with the approximate symmetry, whereas the unconstrained model shows no such degeneracy.}
\label{fig:loss_landscape}
\end{figure}

To probe the effect of the soft symmetry-inducing bias on loss geometry, we perturb the trained model parameters 
along the leading Hessian eigenvector $\nu_1$, corresponding to the direction of largest curvature. 
In the constrained model, this reveals a direction along which the loss varies only weakly, producing an 
approximately degenerate, low-curvature region as depicted in Figure~\ref{fig:loss_landscape}. 
We refer to this as a pseudo-Goldstone mode, in analogy with Goldstone modes in quantum field theory~\cite{peskin1995introduction}. 
The unconstrained model exhibits no comparable direction.

Formally, let $f_{\boldsymbol{\theta}}(x)$ denote the network output for input $x$, and let $g \cdot x$ 
denote the Lorentz-transformed input under $g \in G$. For small perturbations $\boldsymbol{\theta} \to \boldsymbol{\theta} + \epsilon \, \nu_1$, 
the pseudo-Goldstone mode is characterized by approximate invariance along $\nu_1$:
\begin{equation}
f_{\boldsymbol{\theta} + \epsilon \nu_1}(x) \approx f_{\boldsymbol{\theta} + \epsilon \nu_1}(g \cdot x), 
\qquad \forall \, g \in G, \, \epsilon \ll 1.
\end{equation}
Defining the ratio
\begin{equation}
R(\epsilon) = \frac{f_{\boldsymbol{\theta} + \epsilon \nu_1}(x)}{f_{\boldsymbol{\theta} + \epsilon \nu_1}(g \cdot x)},
\end{equation}
we observe that $R(\epsilon) \approx 1$ in the constrained model.
Together, these observations show that a soft symmetry-inducing bias reshapes the loss geometry to introduce low-curvature, symmetry-aligned pseudo-Goldstone directions. While we explicitly demonstrate this effect only along $\nu_1$, similar approximate degeneracies are expected to extend to subleading modes. We focus on the leading direction as it corresponds to the most sensitive local deformation of the loss and decision boundary.

\begin{figure}[h!]
\centering
\includegraphics[width=0.49\textwidth]{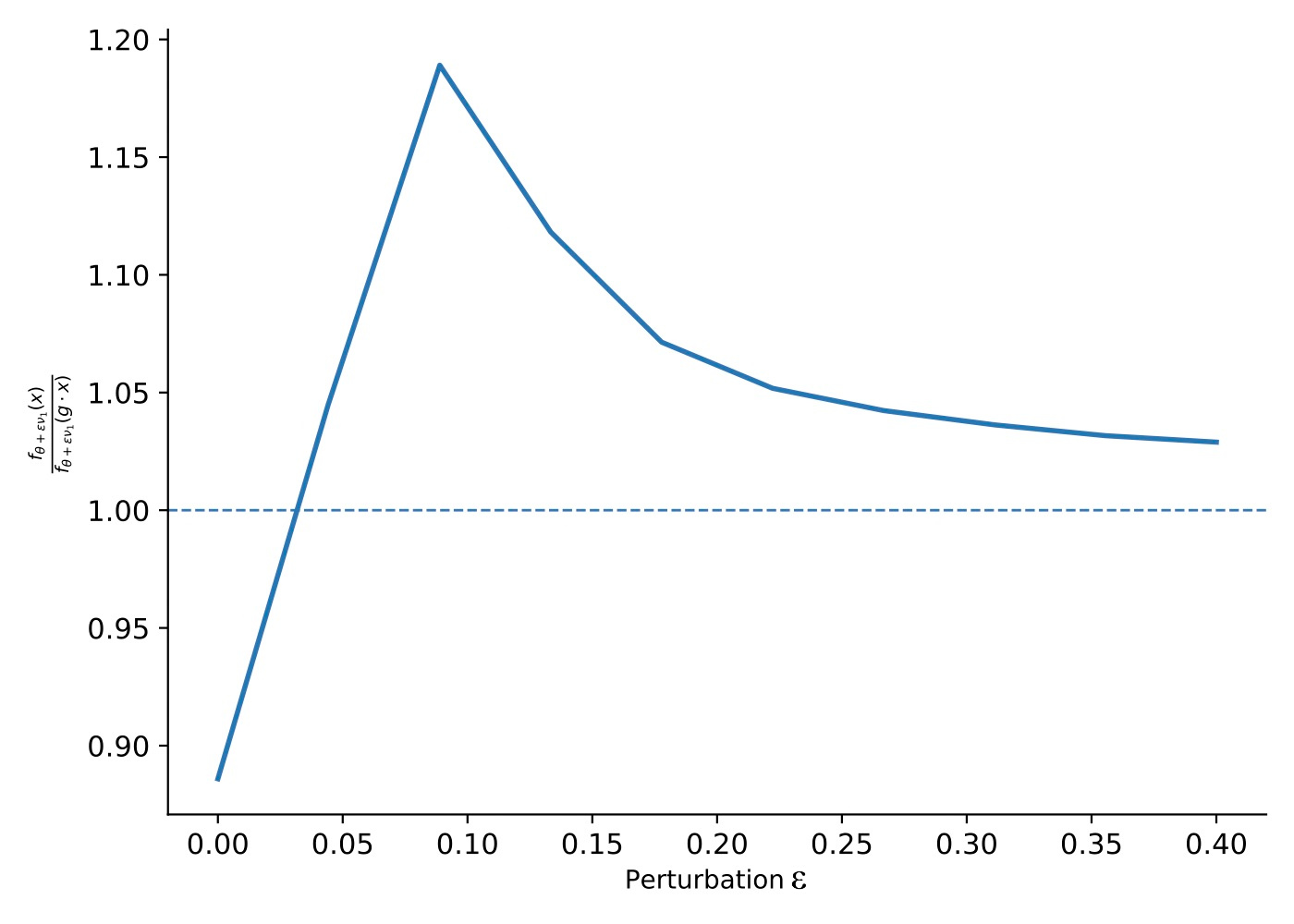}
\caption{Ratio $R(\epsilon)$ as a function of perturbations $\epsilon$ along the leading Hessian eigenvector $\nu_1$ for the constrained model.  The ratio approaches unity, indicating an approximate pseudo-Goldstone direction.}

\label{fig:lorentz_invariance}
\end{figure}

\subsection{Robustness to Input Perturbations}

To test the relationship between loss geometry and robustness, we evaluate model stability under controlled perturbations to the input features of both gluon and W boson jet classes. Each jet constituent's transverse momentum $p_T$ is progressively smeared according to,
\begin{equation}
p_T^\prime = p_T \, \big( 1 + \sigma \, \epsilon \big), \qquad \epsilon \sim \mathcal{N}(0,1),
\end{equation}
where $\sigma$ controls the magnitude of the perturbation. Both constrained and unconstrained models are evaluated on these modified inputs, and their classification performance is quantified using the ROC-AUC metric as a function of $\sigma$. This procedure provides a direct measure of the sensitivity of the model to deviations in input variables. The constrained model exhibits reduced sensitivity across the perturbation range, supporting the interpretation that smaller leading Hessian eigenvalues $\lambda_1$ correlate with greater robustness to both parameter and input perturbations.

\begin{figure}[H]
    \centering
    \includegraphics[width=0.45\textwidth]{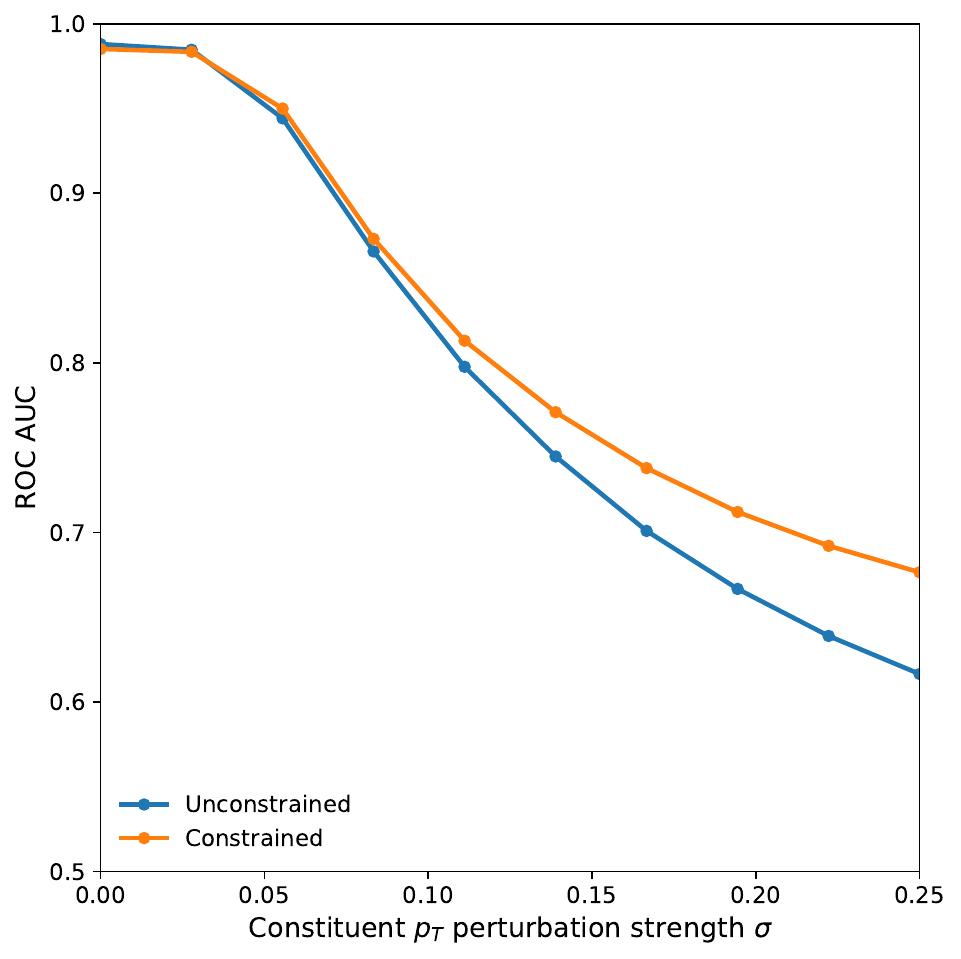}
    \caption{Effect of progressively smearing the transverse momentum ($p_T$) of jet constituents on classification performance. The constrained model exhibits reduced sensitivity to input perturbations compared to the unconstrained baseline.}
    \label{fig:perturbations}
\end{figure}

\subsection{Out-of-Distribution Generalisation}

We evaluate the ability of the models to generalise to out-of-distribution (OOD) data, beyond the training set of W boson and gluon jets. By holding gluons fixed as the common class, we assess whether models have learned transferable abstractions. Two OOD regimes are considered:  

\begin{itemize}
    \item \textbf{Near OOD:} W bosons are replaced by Z boson decays, which are kinematically similar to the training targets.
    \item \textbf{Far OOD:} Top-quark decays are used, which exhibit substantially different kinematics and jet substructure.
\end{itemize}

For each OOD set, we measure classification performance using accuracy and ROC-AUC. These metrics provide a controlled assessment of how well the learned abstractions transfer beyond the training distribution. Across both regimes, the constrained model consistently outperforms the unconstrained baseline. Notably, in the far OOD regime, the unconstrained model achieves an ROC-AUC below 0.5, indicating that it has learned task-specific patterns which fail to generalise. These results demonstrate that the soft inductive bias reshapes the loss geometry to enable more generalisable abstraction learning.

\begin{table}[H]
    \centering
\caption{OOD classification performance in near (Gluon vs Z) and far (Gluon vs Top) regimes. In both cases constrained model outperforms the unconstrained baseline.}
    \vspace{0.3cm}
    \label{tab:performance-summary}
    \begin{tabular}{l l c c}
        \hline
        Task & Model & Accuracy & ROC-AUC \\
        \hline
        \multirow{2}{*}{Gluon vs Z} 
            & Unconstrained & $0.7656 \pm 0.0064$ & $0.9364 \pm 0.0025$ \\
            & Constrained   & $0.7835 \pm 0.0186$ & $0.9429 \pm 0.0078$ \\
        \multirow{2}{*}{Gluon vs Top} 
            & Unconstrained & $0.5258 \pm 0.0028$ & $0.4177 \pm 0.0495$ \\
            & Constrained   & $0.5407 \pm 0.0003$ & $0.5323 \pm 0.0124$ \\
        \hline
    \end{tabular}
\end{table}

\begin{figure}[H]
    \centering

    \begin{minipage}{0.48\textwidth}
        \centering
        \includegraphics[width=\textwidth]{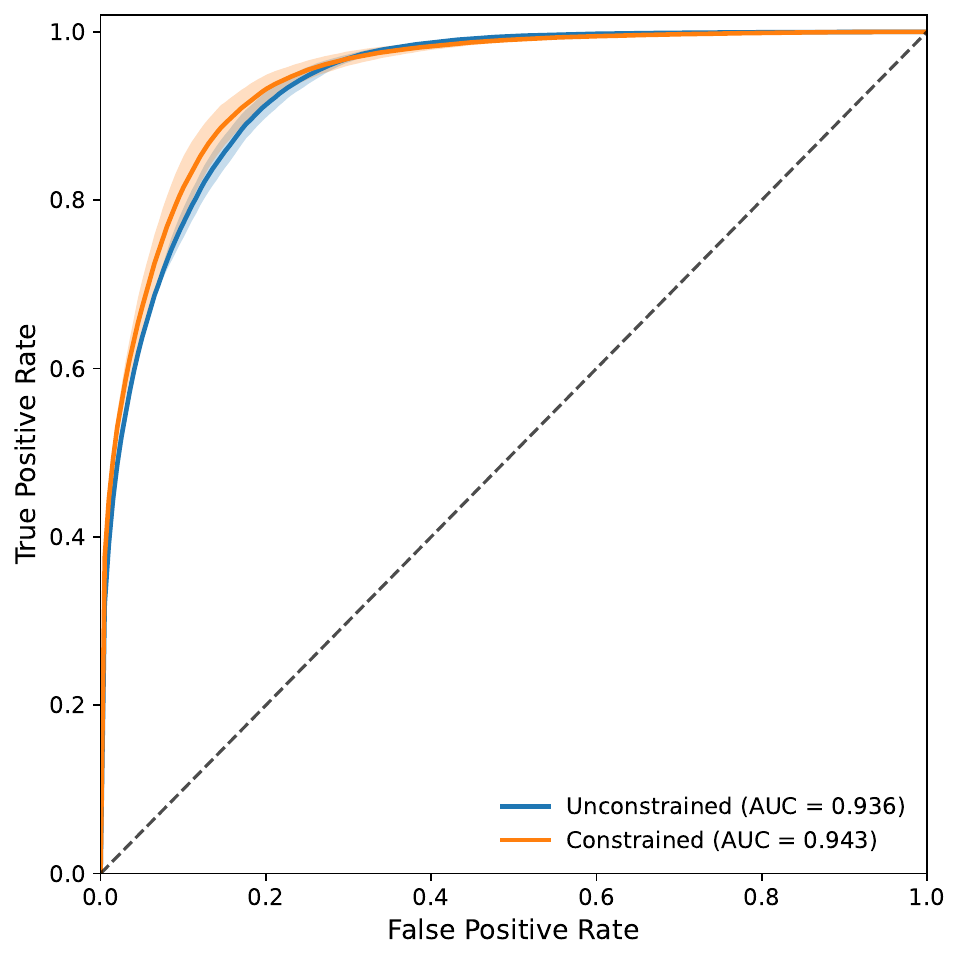}
        (a) Gluon vs Z (near off-manifold)
    \end{minipage}
    \hfill
    \begin{minipage}{0.48\textwidth}
        \centering
        \includegraphics[width=\textwidth]{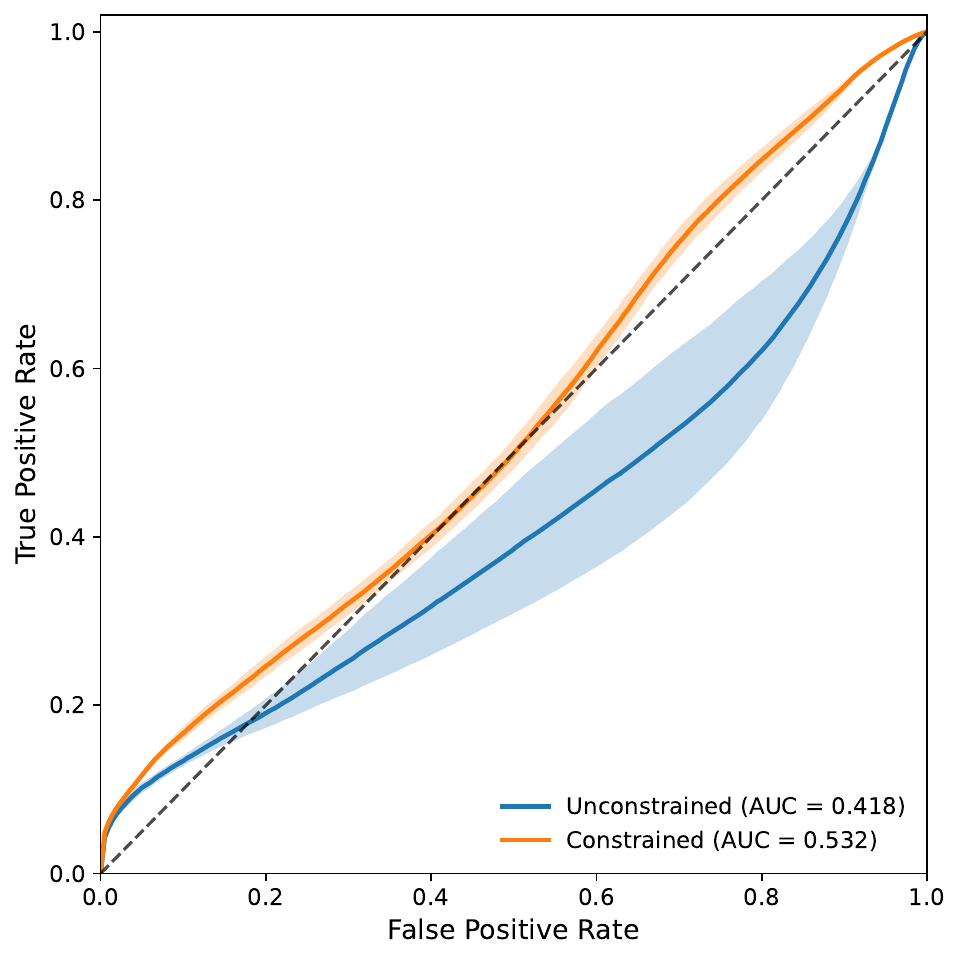}
        (b) Gluon vs Top (far off-manifold)
    \end{minipage}

    \caption{ROC curves for out-of-distribution generalisation. Near OOD test with Z bosons (left). Far OOD test with top quark jets (right).}
    \label{fig:roc-ood}
\end{figure}

\subsection{Distillation and Compressibility}

\begin{figure}[H]
    \centering
    \includegraphics[width=0.7\textwidth]{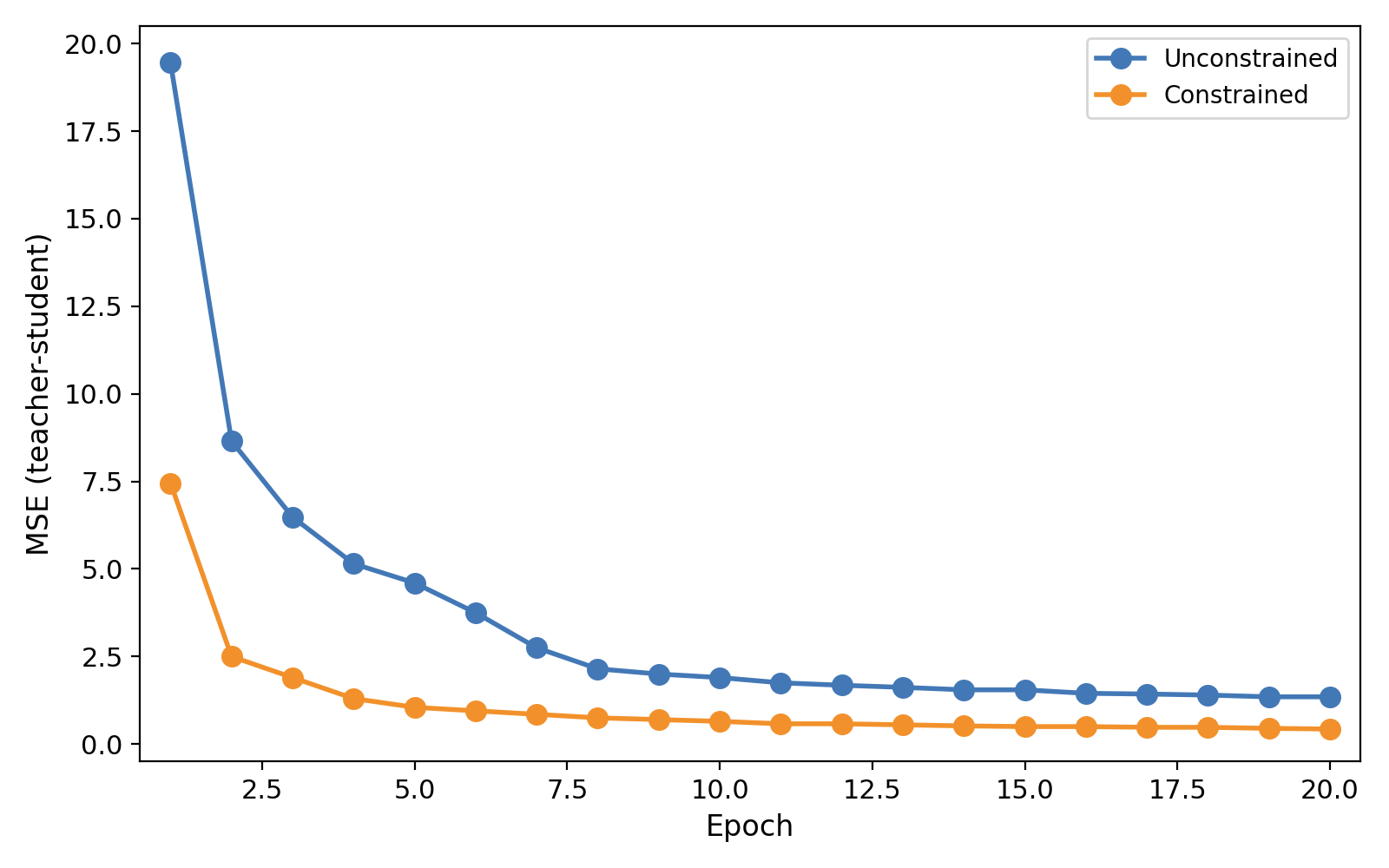}
    \caption{Distillation performance of student networks trained to match unconstrained and constrained teacher models. The constrained teacher yields lower final MSE and faster convergence, indicating reduced functional complexity.}
    \label{fig:distillation}
\end{figure}

The final experiment quantifies the functional complexity through compressibility. We do this by distilling each trained teacher into a fixed, lower-capacity student network. Figure~\ref{fig:distillation} shows the MSE between teacher and student outputs as a function of distillation epochs. The constrained teacher consistently distils faster and reaches a lower final MSE error than the unconstrained baseline. Since the student architecture and optimisation procedure are identical in both cases, this difference directly reflects the reduced functional complexity of the constrained solution.

\section{Discussion and Conclusions}

While data compression is a necessary ingredient for learning effective abstractions, this work demonstrates that simplicity biases present during optimisation also play a crucial role. Both theory and empirical evidence suggest that solutions of lower functional complexity tend to generalise better and be more robust. This is partly because such solutions occupy a larger volume in parameter space, which under PAC‑Bayes intuitions leads to a tighter bound on the expected generalisation error~\cite{pac}. To make these solutions learnable, inductive biases are used to reshape the loss geometry. Here, we study a particular symmetry-constraining soft inductive bias applied to a jet classification task and provide a mechanistic description of its effect on the learning process: 

\begin{itemize}
    \item \textbf{Loss geometry:}  
    A soft symmetry constraint reshapes the loss geometry by introducing low-curvature directions aligned with the underlying symmetry, which appear as approximate degeneracies (pseudo-Goldstone modes) in parameter space.

    \item \textbf{Functional complexity:}  
    These modes reduce the functional complexity of the solution, as quantified by Hessian-based measures and by increased compressibility.

    \item \textbf{Generalisation, robustness, and distillation:}
    Simpler solutions retain in-distribution performance while exhibiting improved robustness to input perturbations, stronger out-of-distribution generalisation, and more efficient distillation into lower-capacity models.
\end{itemize}

It is important to emphasise that parameter count alone is not a sufficient proxy for functional complexity of a learnt solution; meaningful measures require Hessian-based metrics and assessment of compressibility. More broadly, our findings suggest a practical strategy for future model design. High-capacity models, when guided by appropriate inductive biases, can navigate a high-dimensional loss geometry toward low-complexity solutions that are both highly generalisable and robust. Owing to their reduced functional complexity, the abstractions learned by the constrained models can be distilled into smaller networks more efficiently. Smaller models have sufficient capacity to represent these functions. However, they are unable to explore them during optimisation, future work will explore leveraging this approach to train and deploy powerful models in resource and latency constrained HEP environments.

\section*{Acknowledgments}
This work is supported by the Eric \& Wendy Schmidt Fund for Strategic Innovation through the CERN Next Generation Triggers project (grant agreement SIF-2023-004).

Maciej Glowacki acknowledges the contributions of all members of the Next Generation Triggers project for enabling this research.

% ...

\bibliographystyle{unsrt}  
\bibliography{references}  

\end{document}